\documentclass[10pt,twocolumn,letterpaper]{article}

\usepackage{cvpr}              
\usepackage[accsupp]{axessibility}

\usepackage{graphicx}
\usepackage{amsmath}
\usepackage{amssymb}
\usepackage{booktabs}

\usepackage{multirow}
\usepackage{algorithm}
\usepackage{algorithmic}

\usepackage[capitalize]{cleveref}
\crefname{section}{Sec.}{Secs.}
\Crefname{section}{Section}{Sections}
\Crefname{table}{Table}{Tables}
\crefname{table}{Tab.}{Tabs.}

\newtheorem{definition}{Definition}
\newtheorem{theorem}{Theorem}
\newtheorem{lemma}{Lemma}
\newtheorem{assum}{Assumption}
\newtheorem{coll}{Corollary}

\def\cvprPaperID{1250} 
\def\confName{CVPR}
\def\confYear{2022}

\begin{document}

\title{Differentially Private Federated Learning with Local Regularization and Sparsification}

\author{
Anda Cheng\textsuperscript{\rm 1,2} 
\text{ }\text{ }\text{ }\text{ }\text{ }\text{ }
Peisong Wang\textsuperscript{\rm 1}
\text{ }\text{ } \text{ }\text{ }\text{ }\text{ }
Xi Sheryl Zhang\textsuperscript{\rm 1}
\text{ }\text{ }\text{ }\text{ }\text{ }\text{ }
Jian Cheng\textsuperscript{\rm 1,2}\textsuperscript{\rm}\thanks{Corresponding Author.}
\\
\textsuperscript{\rm 1}Institute of Automation, Chinese Academy of Sciences \\
\textsuperscript{\rm 2}School of Artificial Intelligence, University of Chinese Academy of Sciences\\
{\tt\small 
chenganda2017@ia.ac.cn }
\text{ }\text{ }
{\tt\small sheryl.zhangxi@gmail.com}
\text{ }\text{ }
{\tt\small {\{peisong.wang,jcheng\}}@nlpr.ia.ac.cn}
}

\maketitle

\begin{abstract}

User-level differential privacy (DP) 
provides certifiable privacy guarantees
to the information that is specific to any user’s data in federated learning. 
Existing methods that ensure user-level DP come at the cost of severe accuracy decrease.
In this paper, we study the cause of model performance degradation in federated learning with user-level DP guarantee.
We find the key to solving this issue is to naturally restrict the norm of local updates before executing operations that guarantee DP. 
To this end, we propose two techniques, \textit{Bounded Local Update Regularization} and \textit{Local Update Sparsification}, to increase model quality without sacrificing privacy.
We  provide  theoretical  analysis  on  the  convergence  of our framework and give rigorous privacy guarantees. Extensive experiments show that our framework significantly improves the privacy-utility trade-off over the state-of-the-arts for federated learning with  user-level DP guarantee.

\end{abstract}


\section{Introduction}

Federated learning (FL) \cite{fed} is a promising paradigm of distributed machine learning with a wide range of applications \cite{Kairouz2021AdvancesAO,Li2020FederatedLC,FCIL_Dong_CVPR2022}. 
FL enables distributed agents to collaboratively learn a centralized model under the orchestration of the cloud without sharing their local data.
By keeping data usage local, FL
sidesteps the ethical and legal concerns and is advantageous in privacy compared with the traditional centralized learning paradigm. 

However, FL alone does not protect the agents or users from inference attacks that use
the output information. 
Extensive inference attacks demonstrate that
it is feasible to 
infer the subgroup of people with a specific property \cite{Melis2018InferenceAA},
identify individuals \cite{mia}, 
or even infer completion of social security numbers \cite{Carlini2019TheSS}, 
with high confidence from a trained model.

To solve these issues,
differential privacy (DP) \cite{dp}  has been applied to FL in order to protect either each instance in the dataset of any agent (instance-level DP)
\cite{Sun2021FederatedMD, Hu2021FederatedLW, LDPFL}, 
or the whole data of any agent (user-level DP) \cite{dpfedavg,Geyer2017DifferentiallyPF, Kairouz2021TheDD}.
These two DP definitions on different levels are suitable for different situations.
For example, when several banks aim to train a fraud detection model via FL, instance-level DP is more suitable to protect any individual records of any bank from being identified.
In another situation, when a smartphone app attempts  to learn a face recognition model from users’ face images, it is more appropriate to apply  user-level DP to protect each user as a unit. 

Existing methods that ensure  user-level DP \cite{dpfedavg, Geyer2017DifferentiallyPF, Kairouz2021TheDD} are predominantly built upon Gaussian mechanism
which is a Gaussian noise perturbation-based technique.
Unfortunately, directly applying the Gaussian mechanism to ensure strong  user-level DP in FL drastically degrades the utility of the resulted models.
Specifically, the Gaussian mechanism requires to  clip the $l_2$ magnitude of local updates to a sensitivity threshold $S$ and adding noise proportional to $S$ to 
the high dimensional local updates.
These two steps lead to either large bias (when $S$ is small) or large variance (when $S$ is large), which slows down the
convergence and damages the performance of the global model \cite{Zhu2020VotingbasedAF}.
However, existing methods \cite{dpfedavg, Geyer2017DifferentiallyPF, Kairouz2021TheDD}
do not explicitly involve interaction between the operations for ensuring DP and the learning process of FL, which makes the learning process hard to adapt to the clipping and noise perturbation operations, thereby leading to utility degradation of the learned models.

To address the above issues, in this paper, we propose two techniques to improve the model utility in FL with  user-level DP guarantees.
Our motivation is to naturally reduce the $l_2$ norm of local updates before clipping, thereby making the local updates more adaptive to the clipping operation.
First, we propose \textit{Bounded Local Update Regularization (BLUR)}. It introduces a regularization term to the agent's local objective function and explicitly regularizes the $l_2$ norm of local updates to be bounded.
As a  result, the $l_2$ norm of local updates could be naturally smaller than $S$, thereby decreasing the impact of clipping operation.
Then we propose \textit{Local Update Sparsification (LUS)} to further reduce the magnitude of local updates.
Before clipping, it zeros out some update values that have little effect on the performance of the local model, thereby reducing the norm of local updates without damaging the accuracy of the local model.

Our contributions can be summarized as follows:
\begin{itemize}
    \item We propose two techniques to improve the model utility with  user-level DP guarantee in federated learning.

    \item We provide theoretical analysis on the convergence of our framework and give rigorous privacy guarantees.
    
    \item Extensive experiments validate the effectiveness and advantages of the proposed methods.
    
    
\end{itemize}

\section{Related Work}
The concept of  user-level differential privacy in federated learning was introduced by \cite{dpfedavg}.
They propose DP-FedAvg to train models for next-word prediction in a mobile keyboard meanwhile ensuring user-level DP guarantee
by employing Gaussian mechanism and composing privacy guarantees via moment accountant.
The following work \cite{Kairouz2021TheDD,Thakkar2019adaclip} ensures  user-level DP by discretizing the data and adding discrete Gaussian noise before performing secure aggregation.
They also provide a novel privacy analysis for sums of discrete Gaussians. 
Both of the above methods ensure  user-level DP via noise perturbation-based method, which requires to clip norm of model update or data and add noise to the clipped vectors.
Nevertheless, the clipping and noise perturbation steps inevitably interfere with the performance of the resulting model.
Different from the aforementioned methods, a recent study \cite{Zhu2020VotingbasedAF} proposed AE-DPFL which ensured user-level DP by a voting-based mechanism with secure  aggregation. AE-DPLF does not need to clip the model or data, thereby  relieving the accuracy degradation issue. 
However, the AE-DPFL framework assumes that unlabeled data from the global distribution is available to the server, which is very hard to satisfy in practical applications. 
Our work follows the paradigm of the noise perturbation methods but we aim to improve the training process by naturally bounding the local update norms.

Other works related to  our paper are those employing regularization or sparsification techniques in FL.
Previous works \cite{fedprox} and \cite{Acar2021FederatedLB} also introduce regularization terms into objective function for each device.
Nevertheless, they aim to apply the regularization technique to address the data/device distribution heterogeneity issue in FL, which is different from our goal of bounding the sensitivity of local updates.
Another line of works \cite{Agarwal2018cpSGDCA, Hu2021FederatedLW} also apply the sparsification technique in privacy-preserving FL. 
Both of them focus on ensuring instance-level DP and employing sparsification to reduce communication costs.
On the contrary, our work utilizes the sparsification technique to improve the model utility with  user-level DP guarantee.

\section{Preliminary}

\subsection{Federated Learning (FL)}
Federated learning \cite{fed} is a multi-round protocol between an aggregation server and a set of agents in which agents jointly train a model. 
 Let $\mathcal{P}$ denotes the set of all agents with $|\mathcal{P}|=N$, and $\mathcal{D}_i$ denote the local dataset of client $i \in \mathcal{P}$ with $n_i$ samples. 
 The set $\mathcal{D}=\bigcup_{i \in \mathcal{P}} \mathcal{D}_i$ denotes the full training set. 
 Let $f_i(\mathbf{w}, z)$ denotes  the  loss  function for client $i$  over  a  model $\mathbf{w}$ and a sample $z$, and $f_i\left(\mathbf{w}, \mathcal{D}_i\right)=\frac{1}{n_{i}} \sum_{z \in \mathcal{D}_i} f_i(\mathbf{w}, z)$ denotes the empirical loss over a model $\mathbf{w}$ and a dataset $\mathcal{D}_i$.
Without causing ambiguity, we also denote the local loss function as $f_i(\mathbf{w})$ in the following.
In FL, agents try  to jointly train a model that minimizes  the weighted average of local loss functions:
\begin{align}
\min_{\mathbf{w} \in \mathbb{R}^d} \{ f(\mathbf{w}, \mathcal{D})=\sum_{i \in \mathcal{P}} \frac{n_{i}}{n} f_i\left(\mathbf{w}, \mathcal{D}_i\right) \}
\end{align}
where $n \! =\! \sum_{i \in \mathcal{P}} n_{i}$ is the total dataset size of all agents.
 To solve this optimization task, the widely used FedAvg protocol executes the following two steps in the communication round $t$:
 \begin{itemize}
	\item Local updating. The server samples a set of agents $\mathcal{P}_t$. Each agent $i \in \mathcal{P}_t$ downloads the global model $\mathbf{w}^{t-1}$ from server, then performs local training on local dataset by executing $\mathbf{w}_{i}^{t, q} \leftarrow \mathbf{w}_{i}^{t, q-1}-\eta_l \nabla_{\mathbf{w}} f_i\left(\mathbf{w}_{i}^{t, q-1}, \mathcal{D}_i\right)$ for $Q$ steps with $\mathbf{w}^{t, 0}_{i}$  initialized as $\mathbf{w}^{t-1}$. Finally, each agent uploads the model update $\Delta_{i}^{t}=\mathbf{w}^{t,Q}_{i}-\mathbf{w}^{t,0}_{i}$ to server.
	
	\item Model aggregation.  The server receives model updates $\{ \Delta_{i}^{t} | i \in \mathcal{P}_t \}$ from participants and aggregates them to update global model by $\mathbf{w}^{t} \leftarrow \mathbf{w}^{t-1} + \eta_g \sum_{i \in \mathcal{P}_{t}} \frac{n_{i}}{n_{\mathcal{P}_{t}}} {\Delta}_{i}^{t}$.
\end{itemize}

\subsection{Differential Privacy (DP)}

Differential privacy \cite{dp} is a formal notion of privacy that provides provable guarantees against the identification of individuals in a private set.
We denote ${D} \simeq {D}^{\prime}$ as a pair of \textit{adjacent datasets}, which means that ${D}^{\prime}$ can be obtained from ${D}$ by changing only one record.
\begin{definition} \textbf{Differential Privacy.}
A randomized algorithm $\mathcal{M}$ satisfies $(\epsilon, \delta)$-differential privacy if for any adjacent datasets ${D} \simeq {D}^{\prime} $ for any subset of outputs $\mathcal{S} \subseteq Range(\mathcal{M}) $  it holds that $ \operatorname{Pr}[\mathcal{M}({D}) \in S] \leq e^{\epsilon} \operatorname{Pr}\left[\mathcal{M}\left({D}^{\prime}\right) \in S\right]+\delta $.
\end{definition}
The definition indicates that one could not distinguish between ${D}$ and ${D}^{\prime}$ by observing the output of $\mathcal{M}$, thereby protecting individuals in ${D}$ from identification.
A simple way to achieve $(\epsilon, \delta)$-DP is to take a vector-valued deterministic function $F$ and inject appropriate Gaussian noise, the scale of which depends on the sensitivity of $F$.
\begin{definition}
\text{\rm ($l_2$ Sensitivity).}
Let $\mathcal{F}$ be a function, the $L_{2}$-sensitivity of $\mathcal{F}$ is defined as $\mathcal{S}=\max _{D \simeq D^{\prime} } \| \mathcal{F}\left(D\right)-$ $\mathcal{F}\left(D^{\prime}\right) \|_{2}$, where the maximization is taken over all pairs of adjacent datasets.
\end{definition}

\begin{lemma}
Let $\mathcal{F}$ be a function, $\delta \in (0,1)$ and $\epsilon \textgreater 0$. For $c> \sqrt{2 \ln (1.25 / \delta)}$,
the Gaussian mechanism
$\mathcal{F}(\mathcal{D})+\mathcal{N}(0, \sigma^2\mathbf{I})$ with parameter  $\sigma \geq c \mathcal{S} / \epsilon$ ensures $(\epsilon, \delta)$-DP.
\end{lemma}


\subsection{Differential Privacy for Federated Learning}

In FL, DP can be defined into
\textit{instance-level DP} and \textit{user-level DP}, depending on how adjacent dataset is defined.
Our work focuses on the latter.


\begin{definition} \textbf{User-level DP}.
When $\mathcal{D}^{\prime}$ is constructed by adding or removing \textbf{one agent} with all its data records.
\end{definition}
DP-FedAvg \cite{dpfedavg} is the first to  guarantee  user-level DP in FL by applying
Gaussian mechanism.
To ensure  user-level DP, before uploading local updates to the server, 
DP-FedAvg
clips the norm of per-agent model update $\Delta_{i}^{t}$ to a threshold $S$
and 
adds scaled Gaussian noise 
to the bounded update, as shown in Alg. \ref{DP-FedAvg}.
Although DP-FedAvg ensures  user-level DP, it severely  harms the utility of the resulted  models.
In this work, we aim to develop a federated learning framework that has little negative impact on the model utility meanwhile ensuring  user-level DP.

\begin{algorithm}[tb]
\caption{DP-FedAvg}
\label{DP-FedAvg}
\textbf{Input}: Agent sampling probability $p\in (0,1]$, clipping threshold $S$, noise scale $\sigma$. \\
\textbf{Output}: Trained model $\mathbf{w}^T $ \\
\textbf{Server} 
\begin{algorithmic}[1] 
\STATE Initialize global model $\mathbf{w}_0$
\FOR{$t=1$ to $T$}
\STATE $\mathcal{P}_t \leftarrow$ Sample agents with probability $p$;
\FOR{$i \in \mathcal{P}_t$ in parallel}
\STATE $ \overline{\Delta}_{i}^{t} = \text{LocalUpdate}\left(\mathbf{w}^{t-1}, i \right) $;
\ENDFOR
\STATE $\mathbf{w}^{t} \leftarrow  \mathbf{w}^{t-1} + \eta_g \frac{1}{|\mathcal{P}_t|} \sum_{i \in \mathcal{P}_t} \overline{\Delta}_{i}^{t}$;
\ENDFOR
\STATE \textbf{return} $\mathbf{w}^{T}$
\end{algorithmic}
\textbf{LocalUpdate} 
\begin{algorithmic}[1] 
\STATE $\mathbf{w}_{i}^{t,0} \leftarrow \text{Download } \mathbf{w}^{t-1}$;
\FOR{$q=1$ to $Q$}
\STATE Sample batch $\mathcal{B} \subseteq \mathcal{D}_i$ ;
\STATE $\mathbf{w}_{i}^{t, q} \!\leftarrow \! \mathbf{w}_{i}^{t, q-1} \!-\!
\eta_l \frac{1}{|\mathcal{B}|} \!\sum_{(x,y)\in \mathcal{B}}\! \nabla_{\!\mathbf{w}} f_i\left(\mathbf{w}_{i}^{t,q\!-\!1}, x,y  \right)$;
\ENDFOR
\STATE $\Delta_{i}^{t}=\mathbf{w}_{i}^{t,Q}-\mathbf{w}_{i}^{t,0}$;
\STATE $\widetilde{\Delta}_{i}^{t} = \Delta_{i}^{t} / \max \left(1, \frac{\| \Delta_{i}^{t} \|_{2}}{S}  \right) $;
\STATE \textbf{return} $ \widetilde{\Delta}_{i}^{t} + \mathcal{N}(0, S^2\sigma^2\mathbf{I}_d/|\mathcal{P}_t|)$
\end{algorithmic}
\end{algorithm}

\section{Methodology}
We start by analyzing the impact of clipping and adding noise operations in the local update.
We denote  $\Delta_{i}^{t}$ as the local update at communication round $t$ from agent $i$ before clipping, denote $\widetilde{\Delta}_{i}^{t}$ as  the local update after clipping but before adding noise, and denote $\overline{\Delta}_{i}^{t}$ as
the local update after clipping and adding noise.
Let $d$ denote the dimension of $ \Delta_{i}^{t}$, then the expected \textit{mean-square error} of the estimate $\overline{\Delta}_{i}^{t}$ can be computed as follows
\begin{align}\label{bias}
\mathbb{E}\left[\frac{1}{d} \left\| \overline{\Delta}_{i}^{t} - \Delta_{i}^{t} \right\|_2^2 \right]  & \leq \frac{1}{d} \left( \mathbb{E}\left[\left\| \widetilde{\Delta}_{i}^{t} - \Delta_{i}^{t} \right\|_2^2 + \left\| \overline{\Delta}_{i}^{t} - \widetilde{\Delta}_{i}^{t} \right\|_2^2 \right]\right)
\notag\\ 
& = \frac{1}{d}  \max\left( 0, \| \Delta_{i}^{t} \| - S \right)^2  + 
\frac{\sigma^2S^2}{|\mathcal{P}_t|} 
\end{align}
The detailed  derivation of Eq. \ref{bias} is provided in the Appendix. Eq. \ref{bias} indicates that $\overline{\Delta}_{i}^{t}$ is a biased estimation of $\Delta_{i}^{t}$.
At the right side of Eq. \ref{bias}, the first and second term reflect the deviation introduced by clipping and adding Gaussian noise, respectively.
To minimize the deviation, we can decrease the right side of the inequality in two ways: 
\begin{itemize}
\item Ensuring $\| \Delta_{i}^{t} \|$ is not greater than $S$ for each $i$ and $t$;
\item Using a smaller clipping threshold $S$.
\end{itemize}
The first way indicates that we should somehow limit the $l_2$ norm of the local update to make it smaller than $S$. Intuitively, if $\| \Delta_{i}^{t} \|$ is large, e.g. $\| \Delta_{i}^{t} \| \gg S$, the clipping operation could lead to much of the update information contained in $\Delta_{i}^{t}$ be dropped and makes the resulted $\widetilde{\Delta}_{i}^{t}$ less informative.
The second way indicates that we can use smaller $S$ to limit the impact of Gaussian noise. Intuitively, this works because the variance of added Gaussian noise is proportional to $S^2$. Using smaller $S$ can directly reduce the perturbation effect of adding noise.
However, when we also consider the first way, we can find that it is difficult to reduce the deviation by only reducing $S$ without considering $\| \Delta_{i}^{t} \|$. Because for the same $\| \Delta_{i}^{t} \|$ that is greater than $S$, only reducing $S$ could
increase $\| \Delta_{i}^{t} \| - S$, which enlarges the negative impact of clipping operation.
This indicates that the key to solve the problem is to \textbf{naturally reduce the norm of local updates} at each communication round.

Based on the above observation, we propose two techniques to improve the utility federated learning with  user-level DP guarantee, termed {\textit{Bounded Local Update Regularization}} and {\textit{Local Update Sparsification}}.
Our motivation is to reduce the norm of local updates by regularizing local models and making the local updates sparse.

\subsection{Bounded Local Update Regularization (BLUR)}

In vanilla FedAvg, each agent trains the local model by optimizing the objective function of 
\begin{align}\label{loss}
\min_{\mathbf{w} \in \mathbb{R}^d} f_i(\mathbf{w} )
\end{align}
which does not impose any constraints on weight updates. However, when we apply the Gaussian mechanism to ensure user-level DP, the $l_2$ norm of weight update must be limited to ensure the sensitivity of weight update smaller than a threshold $S$.
To this end, the $l_2$ norm of weight update should be considered as a constraint in the local optimization.
Let $\mathbf{w}^{t}$ denote the local initial weight at communication round $t$. Then the local optimization should be formulated as
\begin{align}
\min_{ \mathbf{w}  \in \mathbb{R}^d} f_i\left(\mathbf{w} \right)
\text{ }\text{ }\text{ }\text{ } \text{ s.t. } \| \mathbf{w}  - \mathbf{w}^{t} \| \leq S
\end{align}
The above formulation can be converted to an unconstrained optimization by transforming the constraint to a regularization term (BLUR) as
\begin{align}
\label{loss+reg}
\min_{ \mathbf{w}  \in \mathbb{R}^d} \{ {h_i}(\mathbf{w}) \triangleq  f_i(\mathbf{w})
+ \frac{\lambda}{2} R_t(\mathbf{w} ) \} \notag \\
\text{where } R_t(\mathbf{w} )  =  \max \left(0, \| \mathbf{w}  \!-\! \mathbf{w}^{t} \|^2 \!-\! S^2 \right) 
\end{align}
Directly optimizing Eq. \ref{loss} 
may lead to $\| \Delta_{i}^{t} \| \gg S$, in which case applying clipping operation to $\Delta$ 
could result in much of the information in $\Delta_{i}^{t}$ being dropped, thereby impeding the convergence of the local training process. 
On the contrary,  $\Delta_{i}^{t}$ obtained by optimizing Eq. \ref{loss+reg} is more adaptive to the clipping operation as the regularization term in Eq. \ref{loss+reg}
effectively limits the $l_2$ sensitivity of $\Delta$ to be smaller than the clipping threshold $S$.

The effect of BLUR can also be interpreted as an adaptive adjustment to the local learning rate by considering both model update norm and learning step. Without using BLUR, the local update can be expressed as 
\begin{align}\label{unroll6}
\mathbf{w}_{i}^{t, Q} -  \mathbf{w}^{t} =  -\eta_{l} \sum_{q=0}^{Q-1} \mathbf{g}_{i}^{t, q}
\end{align}
where $\mathcal{B}_q$ denotes the local batch of data at the local step $q$ and $\mathbf{g}_{i}^{t, q} \! = \!  \frac{1}{|\mathcal{B}_q|} \!\sum_{(x,y)\in \mathcal{B}_q}\! \nabla\!  f_i\! \left(\mathbf{w}_{i}^{t,q\!-\!1}, x,y  \right)$ with  $\mathbb{E}\!\left[\mathbf{g}_{i}^{t, q}\right]\!=\!\nabla f_{i}\left(\mathbf{w}_{i}^{t, q}\right)$. 
The result in Eq. \ref{unroll6} can be easily obtained by unrolling the update step of DP-FedAvg (line of LocalUpdate in Alg. \ref{DP-FedAvg}).
While applying BLUR, the local model  is updated by optimizing Eq. \ref{loss+reg}. 
as 
\begin{align}\label{update}
\mathbf{w}_{i}^{t, q} \leftarrow \mathbf{w}_{i}^{t, q-1}-
\eta_l \frac{1}{|\mathcal{B}|} \!\sum_{(x,y)\in \mathcal{B}}\! \nabla {h_i}\left(\mathbf{w}_{i}^{t, q-1}, x,y  \right)
\end{align}

\begin{lemma}\label{the1}
Suppose at communication round $t$, the local model on agent $i$ is updated by repeating Eq. \ref{update} with $\lambda \textless \frac{1}{\eta_l} $ for $Q$ iterations. Then we have the final local update 
\begin{align}\label{unroll}
\mathbf{w}_{i}^{t, Q} -  \mathbf{w}^{t} =  -\eta_{l} \sum_{q=0}^{Q-1} \gamma_{i}^{t, q} \mathbf{g}_{i}^{t, q} 
\end{align}
where 
$
\gamma_{i}^{t, q} =\left\{\begin{array}{ll}
\! (1-\lambda\eta_l)^q, \!
& \text{if } \| \mathbf{w}_{i}^{t, q} -  \mathbf{w}^{t} \| \textgreater S
\\
1, 
& \text{otherwise }
\end{array}\right.
$
\end{lemma}

Lemma \ref{the1} shows that BLUR introduces an adaptive discount factor $\gamma_{i}^{t, q}$ to the local learning rate.  
At the local step $q$, if the norm of the current update is larger than $S$, the learning rate at this step would be discounted by  $(1-\lambda\eta_l)^q$ to restrict the impact of this update step. On the contrary, if the norm of the current update is smaller than $S$, the effect of this step would not be limited. More concretely, the  training  process is forced to the local optimal that lies in the norm-restricted space.

We note that a similar regularization term has been applied in the previous work FedProx \cite{fedprox}. However, an important distinction between FedProx and our BLUR is that we aim to employ the regularization method to bound the sensitivity of the local updates by $S$, while FedProx applies the regularization method to tackle the statistical heterogeneity problem in federated learning.
As a result, the impact of clipping threshold $S$ is taken into account in our BLUR while FedProx does not involve a threshold in the regularization term.

\subsection{Local Update Sparsification (LUS)}

Sparsification is a widely used technique to improve communication efficiency in distributed training \cite{Lin2018DeepGC, Tsuzuku2018VariancebasedGC, Sattler2019SparseBC} or to reduce the model complexity of DNNs \cite{Han2016DeepCC,Li2017PruningFF}.
Inspired by the previous works, we expect to
further reduce the norm of local updates by 
eliminating some parameter updates which can be removed with less impact on model performance.

Suppose in a local update process, the initial model weight is $\mathbf{w_0}$. 
The model weight after local training is  $\mathbf{w}$ and the corresponding update is $\Delta \mathbf{w}$.
Here, we denote the whole model weight vector as $\mathbf{w}$ and denote a specific parameter in the model as $w$.
We can zero out the update of a specific parameter $w$ to 0 by setting $w\leftarrow w_0$ and get the corresponding model weight $ \widetilde{\mathbf{w}}$ and model update 
$\Delta \widetilde{\mathbf{w}}$. By applying the Taylor series on ${f_i}(\widetilde{\mathbf{w}})$, 
we can get the loss value as 
\begin{align}
f_i\left( \widetilde{\mathbf{w}} \right)=
f_i\left(\mathbf{w} \right)
-\frac{\partial f_i\left(\mathbf{w} \right)}{\partial w}(w_0-w)
+o\left(w^{2}\right)
\end{align}
Ignoring the higher-order term, we have
\begin{align}
\displaystyle\left\lvert  f_i\left(\widetilde{\mathbf{w}} \right)-
f_i\left(\mathbf{w} \right) \right\rvert
=\displaystyle\left\lvert\frac{\partial f_i\left(\mathbf{w}\right)}{\partial w}(w_0-w) \right\rvert
\end{align}
We define the utility cost of zeroing out $\Delta w$ as 
\begin{align}
T(\Delta w; \mathbf{w}) \triangleq
\displaystyle\left\lvert\frac{\partial f_i\left(\mathbf{w} \right)}{\partial w}\Delta w \right\rvert
= \displaystyle\left\lvert\frac{\partial f_i\left(\mathbf{w} \right)}{\partial w}(w_0-w) \right\rvert
\end{align}
Large $T(\Delta w; \mathbf{w})$ indicates that zeroing out $\Delta w$ will lead to much utility cost to $\mathbf{w}$, thereby $\Delta w$ should be preserved in $\Delta \mathbf{w}$. 
On the contrary, the updates that have little impact on model performance would be zeroed out.
Suppose there are $J$ layers in the model.
Let $\mathbf{w}_j \in \mathbb{R}^{d_j}$ denote the weight in the $j$-th layer and let $T_s\left(\Delta \mathbf{w}_j\right)$ denote the $s$-th largest value of set  $\{ T\left(\Delta w; \mathbf{w}\right) \mid w\in\mathbf{w}_j \}$. 
To make local update $\Delta \mathbf{w} $ sparse, we define a mask function 
to generate 0-1 mask matrix
for update $\Delta w$ in the $j$-th layer of model $\mathbf{w}$ as 
\begin{align}
\label{ele-mask}
M_{j}(\Delta w;\mathbf{w}, s_j) \triangleq \left\{\begin{array}{ll}
\!1, \!
& \text{if } T\left(\Delta w; \mathbf{w} \right) \geq T_{s}\left(\Delta \mathbf{w}_j\right) \\
0, 
& \text{otherwise }
\end{array}\right.
\end{align}
where $M_j(w;\mathbf{w}, s) \in \mathbb{R}^{d_j}$ is the mask matrix for layer update $\Delta \mathbf{w}_j$.  
Let $M\left(\Delta \mathbf{w} , s \right)$ denotes the mask matrix for model update $\Delta \mathbf{w}$, which is constructed by applying Eq. \ref{ele-mask} to each layer. 
Then the sparsification process can be expressed as 
\begin{align}\label{sp-final}
\Delta \mathbf{\widetilde{w}} \leftarrow M(\Delta \mathbf{w}, s) \circ \Delta \mathbf{{w}}
\end{align}
where $\circ$ denotes Hadamard-product.
After sparsification, for each layer update $\Delta \widetilde{\mathbf{w}}_j$,
$s_j$ update values from $\Delta {\mathbf{w}}_j$ that have largest $T\left(w; \mathbf{w} \right)$ values are preserved and others are zeroed out.
As a result, $\| \Delta \widetilde{\mathbf{w}} \|$ would be consistently smaller than $\| \Delta \mathbf{w} \|$.
By adjusting $s$, we can control the sparsity of local update, thereby adjusting the norm reduction, to improve the utility of uploaded model updates. 


\begin{algorithm}[tb]
\caption{Local Update with BLUR and LUS}
\label{our-local}
\textbf{Input}: Current global model $\mathbf{w}^{t-1}$, clipping threshold $S$, noise scale $\sigma$, regularization factor $\lambda$, number of preserved update values $s$ \\
\textbf{Output}: Local update
\begin{algorithmic}[1] 
\STATE $\mathbf{w}^{t, 0}_{i} \leftarrow \text{Download } \mathbf{w}^{t-1}$;
\FOR{$q=1$ to $Q$}
\STATE Sample batch $\mathcal{B} \subseteq \mathcal{D}_i$ ;
\STATE Update local model $\mathbf{w}_{i}^{t, q} $ using Eq. \ref{update};
\ENDFOR
\STATE $\Delta_{i}^{t}=\mathbf{w}^{t,Q}_{i}-\mathbf{w}^{t,0}_{i}$;
\STATE  Compute mask matrix $M(\Delta_{i}^{t}, s)$ according to Eq. \ref{ele-mask};
\STATE $\hat{\Delta}_{i}^{t} \leftarrow M({\Delta}_{i}^{t} , s) \circ {\Delta}_{i}^{t}$;
\STATE $\widetilde{\Delta}_{i}^{t} = \hat{\Delta}_{i}^{t} / \max \left(1, \frac{\| \hat{\Delta}_{i}^{t} \|_{2}}{S}  \right) $;
\STATE \textbf{return} $ \widetilde{\Delta}_{i}^{t} + \mathcal{N}(0, S^2\sigma^2/|\mathcal{P}_t|)$
\end{algorithmic}
\end{algorithm}

\section{Theoretical Results}
In this section, we give the formal privacy guarantee and rigorous convergence analysis of
our FL framework.

\subsection{Privacy Analysis}

In this subsection, we give the formal privacy guarantee.
Same with DP-FedAvg, our method applies Gaussian mechanism to each agent's local update to ensure DP guarantee.
At each communication round, the privacy guarantee of our method is equal to that of DP-FedAvg, if applying the same noise scale for both methods.
For privacy cost accumulation, the composition theorem can be leveraged to compose the privacy cost at each round. 
In this paper, we make use of the moments accountant \cite{dpdl, rdp} to obtain tighter privacy bounds than previous strong composition theorem \cite{dp}.

Specifically, the moments accountant tracks a bound of the privacy loss random variable.
Given a randomized mechanism $\mathcal{M}$, the privacy loss at output $o \in Range(\mathcal{M})$ is defined as 
$\ell\left(o; \mathcal{M},  \mathcal{D}, \mathcal{ D^{\prime}, \mathbf{aux}} \right) \triangleq \log \frac{\operatorname{Pr}[\mathcal{M}(\mathcal{D}, \mathbf{aux})=o]}{\operatorname{Pr}\left[\mathcal{M}\left( \mathcal{D}^{\prime}, \mathbf{aux}\right)=o \right]}$.
Then, the privacy loss random variable $\mathcal{L}\left(o; \mathcal{M},  \mathcal{D}, \mathcal{ D^{\prime}, \mathbf{aux}} \right)$
is defined by evaluating the privacy loss at the outcome sampled from $\mathcal{M}(\mathcal{D})$.
In our framework, the auxiliary information at round $t$ is the current global weight $\mathbf{w}^{t-1}$. 
The moments accountant are defined as
$\alpha_{\mathcal{M}}(\lambda) \triangleq \max _{ \mathcal{D}, \mathcal{D}^{\prime}, \mathbf{aux}} \log \mathbb{E}\left[\exp \left(\lambda \mathcal{L}\left(\mathcal{M}, \mathcal{D}, \mathcal{D}^{\prime } , \mathbf{aux} \right)\right)\right]$.
According  to the tail bound of moments accountant, $\mathcal{M}$ is $(\epsilon, \delta)$-DP with $ \delta \! =\! \min_{\lambda} \exp \left(\alpha_{\mathcal{M}}(\lambda)-\lambda \varepsilon\right) $. 
Then, for an adaptive mechanism $\mathcal{M}_{1:K} = \mathcal{M}_1, \ldots, \mathcal{M}_K$, according to the composability of moments accountant, the privacy guarantee of $\mathcal{M}_{1:K}$ can be calculated by $\alpha_{\mathcal{M}_{1: K}}(\lambda) \leq \sum_{k=1}^{K} \alpha_{\mathcal{M}_{k}}(\lambda)$.
Based on Theorem 1 in \cite{dpdl}, we obtain the following theorem for privacy cost accumulation of FedAvg with our method Alg. \ref{our-local} as local update method.

\begin{theorem} \label{dp-g}
\text{\rm (Privacy Guarantee).}
Let $P$ denote the number of participant clients in a communication round. There exist constants $c_1$ and $c_2$ so that given 
the number of communication rounds $T$, for any $\epsilon \textless c_1q^2T $, FedAvg that uses Alg. \ref{our-local} as local update method satisfy $(\epsilon, \delta)$  user-level DP for any $\delta \textgreater 0 $, if we choose $\sigma \ge c_2 \frac{P\sqrt{{T\log(1/\delta)}}}{N\epsilon}$.
\end{theorem}

\subsection{Convergence Analysis}

In this subsection, we present the convergence results of our method for general loss functions. Our analysis is based on the following assumptions:
\begin{assum}\label{a1}
\text{\rm ($L$-Lipschitz Continuous Gradient).}
There exists a constant $L \textgreater 0$, such that $\left\|\nabla f_{i}(\mathbf{x})-\nabla f_{i}(\mathbf{y})\right\| \leq L\|\mathbf{x}-\mathbf{y}\|, \forall \mathbf{x}, \mathbf{y} \in \mathbb{R}^{d}, \text { and } i \in \mathcal{P}$.
\end{assum}

\begin{assum}\label{a2}
\text{\rm (Unbiased Local Gradient Estimator).}
For any data sample $z$ from $\mathcal{D}_i$, the local gradient estimator is unbiased, e.g., $ \mathbb{E}\left[ \nabla f_i(\mathbf{w}, z)\right]\! =\! \mathbb{E}\left[\nabla f_i(\mathbf{w} )
\right], \forall \mathbf{w} \!\in \! \mathbb{R}^d$ and $i \!\in \!\mathcal{P}$.
\end{assum}

\begin{assum}\label{a3}
\text{\rm (Bounded Variance)}. 
There exist two constants  $\sigma_{l}\textgreater 0$ and $\sigma_{g}\textgreater 0$ such that for any $\mathbf{w} \in \mathbb{R}^d $ and $i \in  \mathcal{P}$, the variance of each local gradient estimator is bounded by $\mathbb{E}\left[\left\|\nabla f_{i}\left(\mathbf{w}, z\right)-\nabla f_{i}\left(\mathbf{w}\right)\right\|^{2}\right] \leq \sigma_{l}^2$, for any data sample $z$ from $\mathcal{D}_i$, and the global variance of the local gradient of the cost function is bounded by $\left\|\nabla f_{i}\left(\mathbf{w}\right)-\nabla f\left(\mathbf{w}\right)\right\|^{2} \leq \sigma_{g}^{2}$.
\end{assum}

\begin{assum}\label{a4}
\text{\rm (Bounded Gradient). }
The loss function $f_i(\mathbf{w}; z)$ has $G$-bounded gradients, i.e., for any $\mathbf{w}\in \mathbb{R}^d$, $i\in\mathcal{P}$, and any data sample $z$ from $\mathcal{D}_i$, we have $\| \nabla f_i(\mathbf{w}; z) \| \leq G $.
\end{assum}

Based on the above assumptions, we have the following convergence results:

\begin{theorem}\label{conv}
\text{\rm (Convergence of Our Protocol).}
Under Assumptions \ref{a1}-\ref{a4},
the sequence of outputs $\{\mathbf{w}^t\}$ generated by Alg. \ref{DP-FedAvg} with Alg. \ref{our-local} as local update method satisfies:
\begin{align}
& \frac{1}{T} \sum_{t=1}^T
\mathbb{E}\left[\overline{\alpha}^{t}\left\|\nabla f\left(\mathbf{w}^{t}\right)\right\|^{2}\right] \notag \\ & \leq 
\underbrace{
\mathcal{O}\left(\frac{1}{\eta_{g} \eta_{l} QT} 
+ {\eta_{l}^2 Q^2}
+ \frac{\eta_{g} \eta_{l} }{P}\right)}_{\text{From FedAvg}}+
\underbrace{\mathcal{O}\left( \frac{\eta_{g}\sigma^2S^2d}{\eta_lQP^2}\right)}_{\text{From operations for DP}} \notag 
\end{align}
\end{theorem}
\textit{where }
$
\overline{\alpha}^{t}
:=\frac{1}{N} \sum_{i=1}^{N} \min \left(1, \frac{S}{ \eta_l \beta_{i}^{t} \left\|  \sum_{q=0}^{Q-1} \gamma_{i}^{t,q} \mathbf{g}_{i}^{t,q} \right\|} \right)$ with
$\beta_{i}^{t} = \frac{\|  M_i^t \circ \sum_{q=0}^{Q-1} \gamma_{i}^{t,q} \mathbf{g}_{i}^{t,q}  \|}{\| \sum_{q=0}^{Q-1} \gamma_{i}^{t,q} \mathbf{g}_{i}^{t,q}  \|} \notag
$.

The bound of Theorem \ref{conv}, 
contains the first term inherited from standard FedAvg
and the second term introduced by operations for DP guarantees.
Comparing with the convergence rate of DP-FedAvg, 
our method achieves quadratic speedup convergence with respect to $P$ in the second term, while that of DP-FedAvg is linear speedup \cite{zhang2021understanding}.
To analyze the privacy/utility trade-off of our framework, we can replace the $\sigma$ in Theorem \ref{conv} with that from Theorem \ref{dp-g}. To analyze the impact of privacy parameters, let $S=\eta_lQc$ with $c\geq G$ and $\sigma^2$ substituted. We can obtain the following results about privacy/utility trade-off.

\begin{coll}\label{pu}
\text{\rm (Convergence with Privacy Guarantee).}
Under Assumptions \ref{a1}-\ref{a4}, for any clipping threshold $S \geq \eta_lQG$ and $\sigma$ as in Theorem \ref{dp-g}, for any $(\epsilon, \delta)$ satisfying the constraints in Theorem \ref{dp-g}, we have
\begin{align}
&\frac{1}{T} \sum_{t=1}^{T} \mathbb{E}\left[\overline{\alpha}^{t}\left\|\nabla f\left(\mathbf{w}^t\right)\right\|^{2}\right]
\notag \\ &
\leq \underbrace{\mathcal{O}\left( \frac{1}{\eta_{g} \eta_{l} Q T}+\eta_{l}^{2} Q^{2}+\frac{\eta_{g} \eta_{l}}{P}\right)}_{\text {From FedAvg }}
+\underbrace{\mathcal{O}\left(\frac{\eta_{g} \eta_{l} Q T d \ln \left(\frac{1}{\delta}\right)}{N^{2} \epsilon^{2}}\right)}_{\text {From operations for DP }} \notag
\end{align}
and the best rate one can obtain from the above bound is $\tilde{O}\left(\frac{\sqrt{d}}{N \epsilon}\right)$ by optimizing $\eta_l,\eta_g,Q,T$.
\end{coll}

\renewcommand{\arraystretch}{.8}
\begin{table*}[tbhp]
	\centering  
	\begin{tabular}{@{}lccccccccccccc@{}}
		\toprule
		Model & Setting  & DP-FedAvg & AE-DPFL & DDGauss &  Ours \\
		\midrule
		\multirow{4}{*}{
			\begin{tabular}{@{}l@{}}
				CNN-2-Layers\\
		\end{tabular}}
		& $\epsilon=2.0$ & $69.65\pm0.74$ & $71.16\pm0.47$ & $69.35\pm0.61$ & 
		$\mathbf{ 74.48\pm0.52}$  \\
		& $\epsilon=4.0$ & $72.32\pm0.81$ & $74.63\pm0.59$ & $72.16\pm0.76$ & $\mathbf{75.85\pm0.61}$  \\
		& $\epsilon=6.0$ & $74.12\pm0.75$ & $76.25\pm0.42$ & $74.34\pm0.70$ & $\mathbf{77.48\pm0.54}$ \\
		& $\epsilon=8.0$ & $75.36\pm0.64$ & $77.41\pm0.33$ & $75.20\pm0.68$ & $\mathbf{78.09\pm0.46}$ \\
		\midrule
		\multirow{4}{*}{
			\begin{tabular}{@{}l@{}}
				ResNet-18\\
		\end{tabular}}		
		& $\epsilon=2.0$ & $73.52\pm0.53$ & $76.37\pm0.41$ & $73.16\pm0.58$ & 
		$\mathbf{ 78.58\pm0.39}$  \\
		& $\epsilon=4.0$ & $75.51\pm0.60$ & $79.22\pm0.46$ & $75.65\pm0.64$ & $\mathbf{80.29\pm0.47}$  \\
		& $\epsilon=6.0$ & $77.19\pm0.55$ & $80.24\pm0.37$ & $77.64\pm0.61$ & $\mathbf{81.55\pm0.46}$ \\
		& $\epsilon=8.0$ & $78.06\pm0.49$ & $81.33\pm0.46$ & $78.03\pm0.52$ & $\mathbf{82.12\pm0.52}$ \\
		\bottomrule
	\end{tabular}
	\vspace{-0.2cm}
	\caption{Performance comparison under different privacy budgets on EMNIST dataset. A smaller $\epsilon$ indicates a stronger privacy guarantee. } 
	\label{emnist}  
\end{table*}
\vspace{-0.5cm}
\renewcommand{\arraystretch}{.8}
\begin{table*}[tbhp]
	\centering  
	\begin{tabular}{@{}lccccccccccccc@{}}
		\toprule
		Model & Setting & DP-FedAvg & AE-DPFL & DDGauss &  Ours \\
		\midrule
		\multirow{4}{*}{
			\begin{tabular}{@{}l@{}}
				CNN-2-Layers\\
		\end{tabular}}
		& $\alpha=0.1$  & $53.84\pm1.04$ & $55.79\pm0.86$ & $53.55\pm1.12$ & $\mathbf{58.95\pm0.95}$  \\
		
		& $\alpha=1$ & $58.67\pm0.85$ & $60.00\pm0.57$ & $58.28\pm0.96$ & $\mathbf{63.74\pm0.70}$  \\
		
		& $\alpha=10$ & $62.25\pm0.71$ & $63.93\pm0.45$ & $62.43\pm0.77$ &  $\mathbf{65.34\pm0.52}$ \\
		
		& $\alpha=100$ & $63.73\pm0.64$ & $64.51\pm0.32$ & $63.80\pm0.69$ &  $\mathbf{66.05\pm0.45}$ \\
		\midrule
		\multirow{4}{*}{\begin{tabular}{@{}l@{}}ResNet-18\\\end{tabular}}
		
		& $\alpha=0.1$  & $59.73\pm0.96$ & $63.11\pm0.65$ & $59.37\pm1.04$ & $\mathbf{64.50\pm0.88}$  \\
		
		& $\alpha=1$ & $63.49\pm0.81$ & $65.80\pm0.51$ & $63.84\pm0.89$ & $\mathbf{67.27\pm0.62}$  \\
		
		& $\alpha=10$ & $65.64\pm0.69$ & $67.62\pm0.42$ & $65.85\pm0.72$ &  $\mathbf{68.96\pm0.54}$ \\
		
		& $\alpha=100$ & $66.58\pm0.60$ & $68.39\pm0.35$ & $66.74\pm0.63$ &  $\mathbf{69.42\pm0.47}$ \\
		\bottomrule
	\end{tabular}
	\vspace{-0.2cm}
	\caption{Performance comparison given different data settings on CIFAR-10 dataset. A smaller $\alpha$ indicates higher data heterogeneity.  } 
	\label{cifar}  
\end{table*}

\section{Experiment Settings}
In this section, we conduct experiments to illustrate the advantages of DP-FedAvg with BLUR and LUS over the previous arts for FL with user-level DP guarantees.

\paragraph{Baselines.} 
Our method aims at improving the performance of \textbf{DP-FedAvg}\cite{dpfedavg}. 
As a result, we choose DP-FedAvg as our baseline. 
DP-FedAvg ensures user-level DP guarantee by directly employing Gaussian mechanism to the local updates.
To compare with SOTA methods, we also compare our method with previous works \textbf{DDGauss} \cite{Kairouz2021TheDD} and \textbf{AE-DPFL} \cite{Zhu2020VotingbasedAF}.
DDGauss ensures user-level DP by discretizing the data and adding discrete Gaussian noise before performing secure aggregation.
AE-DPFL ensures user-level DP by a private voting mechanism with secure  aggregation.

\paragraph{Datasets and models.}
We evaluate on two datasets: EMNIST and CIFAR-10.
EMNIST is an image dataset with hand-written
digits/letters over 62 classes grouped into 3400 clients by their writer. 
It substantially involves user-level DP with natural client heterogeneity and non-iid data distribution.
CIFAR-10 is also an image dataset with 50K training samples and 10K testing samples over 10 classes. 
For CIFAR-10 dataset, we follow prior works \cite{hsu2019measuring,zhu2021data} to
model non-iid data distributions using a Dirichlet distribution
$\mathbf{Dir}(\alpha)$, in which a smaller $\alpha$ indicates higher data
heterogeneity, as it makes the local distribution more biased.
For both datasets, we conduct experiments on
two models with different number of parameters: CNN-2-Layers model from \cite{adap-fl} and ResNet-18 \cite{res}. 
The model size is about $1.0$M for CNN-2-Layers model and $11.1$M for ResNet-18.

\paragraph{Configuration.}
For EMNIST and CIFAR-10 respectively, we set
the number of rounds $T$ to 1000 and 300,
the default agent selection probability $p$ to $0.04$ and $0.06$,
the mini-batch size to 64 and 50,
the local LR $\eta_l$ to 0.03 and 0.1
For all experiments, the number of local iterations $Q = 30$, server LR $\eta_g=1$.
The privacy parameter $\delta=\frac{1}{N} $.
For a specific $\epsilon$, the clipping threshold $S$ for vanilla DP-FedAvg is decided by grid search from $\{0.01, 0.03, 0.1, 0.3, 1.0\}$. 
We find $S=0.03$ and $S=0.3$ perform best on EMNIST and CIFAR-10, respectively. 
The hyper-parameter of BLUR is the regularization parameter $\lambda$.
The hyper-parameter of LUS is the number of preserved updates $s$. Instead of using $s$, we define and adjust the \textit{sparsity} $c=1-s/d$. A larger $c$ indicates more update values are zerod out.
While using BLUR and/or LUS, the hyper-parameters $\lambda$ and $c$ are chosen by grid search from $\{0.05,0.1,0.2,0.4,0.8 \}$ and $\{0.1,0.3,0.5,0.7,0.9 \}$, respectively.
The default $\lambda$ and $c$ are set to $\lambda=0.4$ and $c=0.7$.

\section{Experimental Results}
\paragraph{Performance under different privacy budgets.}
Table \ref{emnist} shows the test accuracies for different level privacy guarantees on EMNIST. 
Our method consistently outperforms the previous SOTA methods for private FL under different privacy budgets.
Specifically, using BLUR and LUS can improve the accuracy of DP-FedAvg by $3\% \sim 4\%$  and $4\% \sim 5\%$  for CNN-2-Layers and ResNet-18, respectively. 
Comparing with SOTA methods, our method consistently provides
significant improvements.
For instance, on ResNet-18, our method provides gains of $4\% \sim 5\%$ to DDGauss and $1\% \sim 2\%$ to AE-DPFL.
We also observe that the improvement on the larger model (ResNet-18) is relatively greater than that on the smaller model (CNN), which is a favorable advantage as we tend to use a large model to achieve better performance in practice.
Moreover, the improvement for smaller $\epsilon$ is relatively greater than that for larger $\epsilon$. 
For instance, the accuracy improvement over DP-FedAvg is $4.83\%$ for $\epsilon=2$, and $2.73\%$ for $\epsilon=8$ on CNN-2-Layers model.
This is also a merit of our method as we tend to use smaller $\epsilon$ to ensure stronger DP guarantees.

\paragraph{Effectiveness of BLUR.}
We conduct experiments to validate the effectiveness of BLUR.
The experiments are conducted on EMNIST with ResNet-18.
The privacy budget is $\epsilon=6.0$.
To verify the effectiveness of BLUR, 
we study the performance of DP-FedAvg + BLUR with various regularization hyper-parameter $\lambda$ from $\{0, 0.05, 0.1, 0.2, 0.4, 0.6, 0.8 \}$, where $\lambda=0$ indicates the vanilla DP-FedAvg.
As shown in Figure \ref{conv}, using BLUR consistently speeds up the convergence and improves the test accuracy of DP-FedAvg.

\begin{figure}[htbp]
	\centering
	\includegraphics[scale=0.95]{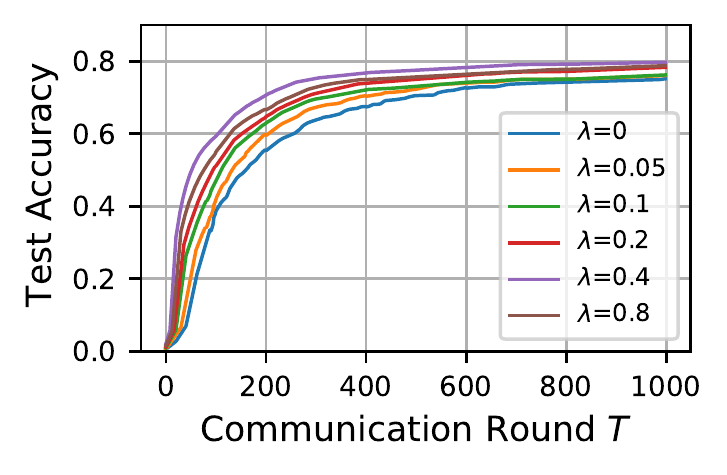}
	\vspace{-0.4cm}
	\caption{Effectiveness of BLUR with various $\lambda$. Vanilla DP-FedAvg is denoted by $\lambda=0$. Using  BLUR can consistently speed up the convergence and improve the test accuracy.}
	\label{conv-fig}
\end{figure}
\vspace{-.5cm}
\renewcommand{\arraystretch}{.8}
\begin{table}[htbp]
	\centering  
	\begin{tabular}{@{}lcccccccccccccc@{}}
		\toprule
		Method & Sparsity & Accuracy (\%) & Gain (\%) \\
		\midrule
		DP-FedAvg & $0.0$ & $76.24$ & $+0.00 $  \\
		\midrule
		\multirow{5}{*}{
			\begin{tabular}{@{}l@{}}
				DP-FedAvg \\ + LUS
		\end{tabular}}
		& $0.1$ & $76.52$ & $+0.28 $ \\
		& $0.3$ & $77.28$ & $+1.04 $ \\
		& $0.5$ & $77.75$ & $+1.51 $ \\
		& $0.7$ & $77.54$ & $+1.30 $ \\
		& $0.9$ & $77.39$ & $+1.15 $ \\
		\midrule
		\multirow{5}{*}{
			\begin{tabular}{@{}l@{}}
				DP-FedAvg \\ + BLUR\\ + LUS
		\end{tabular}}		
		& $0.1$ & $78.28$ & $+2.04$ \\
		& $0.3$ & $79.26$ & $+3.02$ \\
		& $0.5$ & $79.97$ & $+3.73$ \\
		& $0.7$ & $80.32$ & $+4.08$ \\
		& $0.9$ & $80.17$ & $+3.93$ \\		
		\bottomrule
	\end{tabular}
	\vspace{-0.2cm}
	\caption{Effectiveness of LUS with different sparsity. Using LUS consistently improves the accuracy and is synergistic with BLUR. } 
	\label{sp}  
\end{table}
\vspace{-0.2cm}

\begin{figure*}[thbp]
	\centering
	\begin{minipage}[t]{0.5\textwidth}
		\centering
		\includegraphics[scale=0.43]{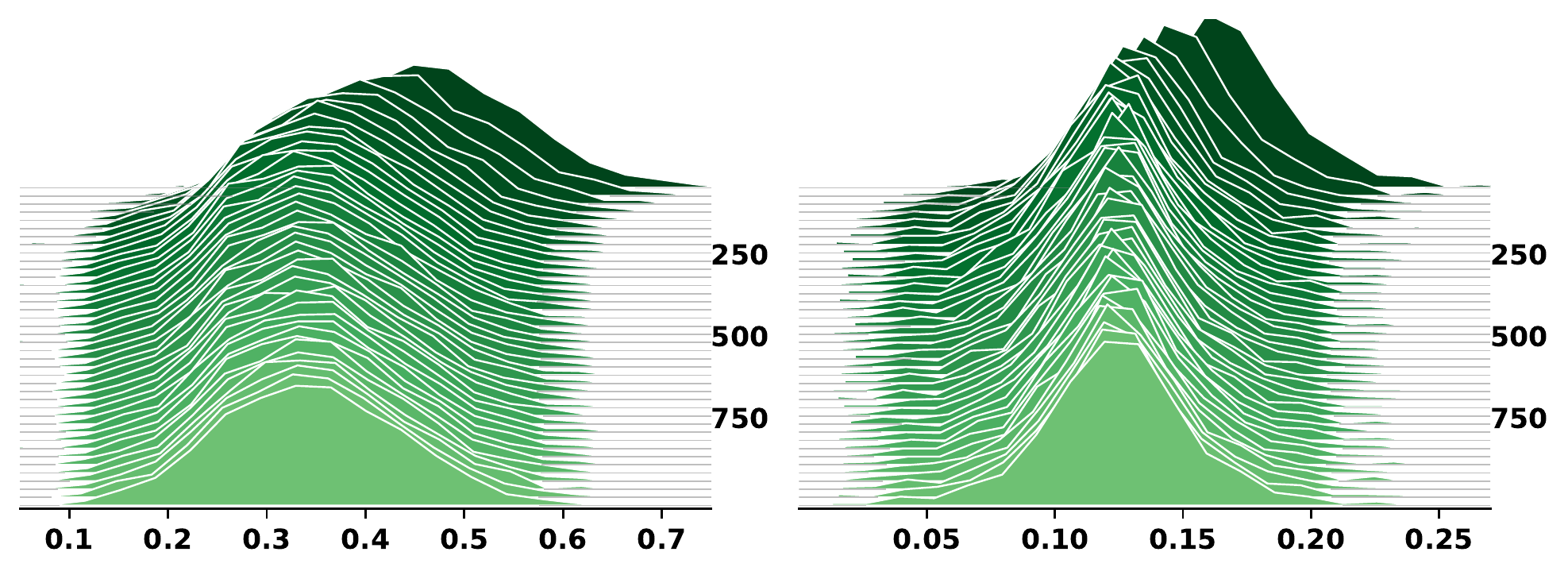}
		\vspace{-0.25cm}
		\caption{Distributions of local update norms (before clipping) at each round from DP-FedAvg (left) and ours (right). The y-axis and x-axis denote communication rounds and local update norms, respectively.}
		\label{norm}
	\end{minipage}
	\hspace{0.2cm}
	\begin{minipage}[t]{0.23\textwidth}
		\includegraphics[scale=0.67]{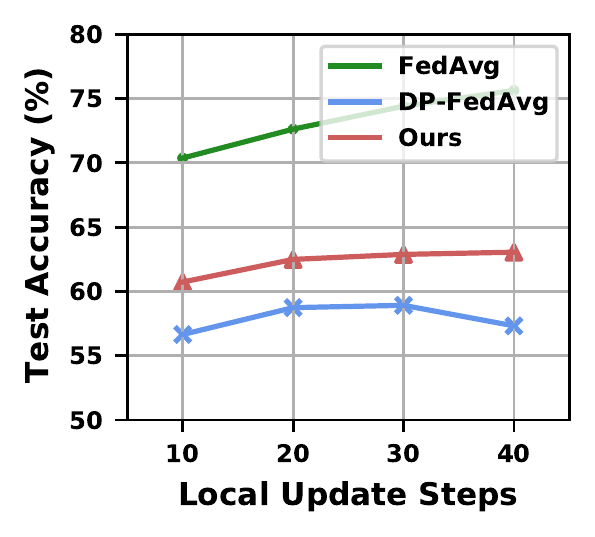}
    	\vspace{-0.64cm}
		\caption{Impact of local update steps on CIFAR-10.}
		\label{Q}
	\end{minipage}
	\hspace{0.3cm}
	\begin{minipage}[t]{0.22\textwidth}
		\includegraphics[scale=0.67]{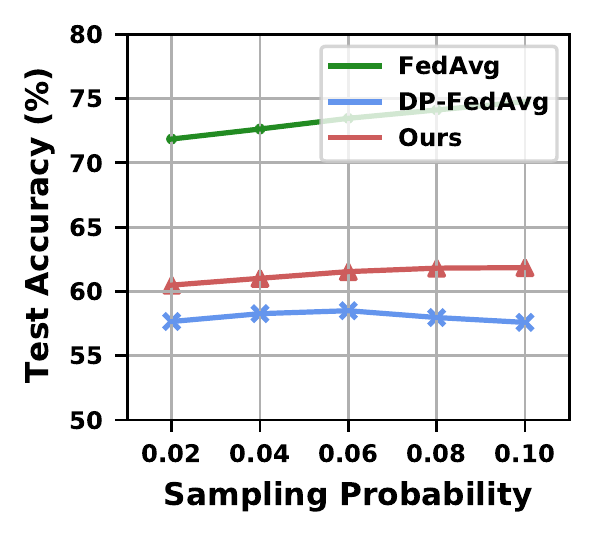}
	    \vspace{-0.64cm}
		\caption{Impact of active agent numbers on CIFAR-10.}
		\label{p}
	\end{minipage}
\vspace{-0.4cm}
\end{figure*}

\paragraph{Effectiveness of LUS.}
To validate the effectiveness of LUS, we conduct 
experiments on DP-FedAvg + LUS and DP-FedAvg + BLUR + LUS with various sparsity $c$ from $\{0,  0.1, 0.3, 0.5, 0.7, 0.9 \}$, where sparsity=0 indicates not using LUS. 
From Table \ref{sp}, we observe that when  equipping DP-FedAvg with LUR solely, the performance of DP-FedAvg are improved by about $0.58\% \sim 1.51\%$. 
However, compared with DP-FedAvg + BLUR, DP-FedAvg + BLUR + LUS obtains more performance gains by at most $2.07\%$, which indicates that the effectiveness of LUS can be boosted while cooperating with BLUR, and certifies that
the effects of BLUR and LUS are synergistic.

\paragraph{Effects of bounding local update norms.}
To verify the effects of our method on bounding the norm of local updates, we show in Figure \ref{norm} the distributions of local updates norm before clipping in each communication round. 
The clipping bound is set to be 0.1 for both DP-FedAvg and our method. 
In contrast to DP-FedAvg, the clipping operation distorts less information in our framework, witnessed by a much smaller difference in the norm of local updates and the clipping threshold, which is smaller than 0.1 in most cases.
Moreover, the local updates used in our method exhibit much less variance compared with DP-FedAvg.
This is consistent with our motivation of making the local updates more adaptive to clipping by naturally reducing the norm of local updates before clipping.

\paragraph{Impacts of data heterogeneity.}
We explore different data heterogeneity by changing $\alpha$ for Dirichlet distribution in Table \ref{cifar}.
We observe that our method consistently outperforms other baselines
for different data heterogeneity.
Moreover, using BLUR and LUS can lead to more accuracy gain when data heterogeneity is higher.
For example, the accuracy gain is $5.11\%$ for $\alpha=0.1$, and $2.32\%$ for $\alpha=100$ on CNN-2-Layers.
The reason for this could be that when data heterogeneity is higher, the local data distribution is more biased to the global distribution, leading to larger norm of local updates. Therefore, the clipped local updates are more biased to the original local updates. Employing BLUR and LUS can mitigate this by bounding the norm of local updates.

\paragraph{Impacts of communication frequency.}
We explore different local updating steps $Q$ on CIFAR-10, so that a larger $Q$ means longer communication delays before the global communication. Results in Figure \ref{Q} indicates that our approach is robust against different levels of communication
delays while DP-FedAvg leads to performance degradation when $Q$ is large, e.g. $Q=40$. 
This is because that updating local models for more steps makes the updated local models more far away from the global model, leading to larger norm of local updates.
On the contrary, our method can effectively limit the norm of local updates, thereby reducing the accuracy drop caused by clipping.

\paragraph{Impacts of active agents.}
We explore different agent sampling probability $p$ on CIFAR-10.
Using larger $p$ means more agents participate in each round of communication but also requires more noise injection to the local updates according to Theorem \ref{dp-g}.
Results in Figure \ref{p} indicates that equipping DP-FedAvg with BLUR and LUS makes it more robust against different levels of agent sampling rates.

\section{Conclusion}
We study the cause of model utility degradation in federated learning with DP and find the key is to naturally bound the local update norms before clipping.
We then propose local regularization and sparsification methods to solve the problem.
We  provide  theoretical  analysis  on  the  convergence and privacy of our framework.
Experiments show that our framework significantly improves model utility over SOTA for federated learning with DP guarantee.

\section*{Acknowledgments}
This work was supported in part by the
National Key Research and Development Program of China (No. 2020AAA0103402),
the Strategic Priority Research Program
of Chinese Academy of Sciences (No. XDA27040300
and
No. XDB32050200),
the National Natural Science Foundation of China (No.62106267).


{\small
\bibliographystyle{ieee_fullname}
\bibliography{egbib}
}

\clearpage

\end{document}